\documentclass{article}
\usepackage{spconf,amsmath,graphicx}
\usepackage{amssymb}
\usepackage{algorithm}
\usepackage[noend]{algpseudocode}
\usepackage{booktabs}
\usepackage{multirow}
\usepackage{xcolor}
\usepackage{url}
\usepackage[hang,flushmargin]{footmisc}
\usepackage{tablefootnote}

\usepackage{subcaption}

\usepackage{enumitem}

\renewcommand{\thefootnote}{\fnsymbol{footnote}}

\title{Adaptive Knowledge Distillation between Text and Speech\\ Pre-trained Models}
\name{\begin{tabular}{c} 
Jinjie Ni$^{1,2,*}$, Yukun Ma$^{1}$, Wen Wang$^{1}$, Qian Chen$^{1}$, Dianwen Ng$^{1,2}$, Han Lei$^{2}$, \\
Trung Hieu Nguyen$^{1}$, Chong Zhang$^{1}$, Bin Ma$^{1}$, Erik Cambria$^{2}$
\end{tabular}}

\address{$^1$Alibaba Group \\
 $^2$School of Computer Science and Engineering, Nanyang Technological University}
\definecolor{green1}{rgb}{0.2, 0.66, 0.32}
\definecolor{YELLOW}{rgb}{0.83, 0.61, 0.18}

\newcommand\blfootnote[1]{%
  \begingroup
  \renewcommand\thefootnote{}\footnote{#1}%
  \addtocounter{footnote}{-1}%
  \endgroup
 }

\begin{document}

%
\maketitle

\addtocounter{footnote}{1}
\blfootnote{$^*$Work done during an internship at Speech Lab, Alibaba Group, Singapore.}

\addtocounter{footnote}{1}
\blfootnote{$^*$Supported by the Agency for Science, Technology and Research (A*STAR) under its AME Programmatic Funding Scheme (Project \#A18A2b0046).}

\vspace{-2em}

\begin{abstract}
Learning on a massive amount of speech corpus leads to the recent success of many self-supervised speech models. With knowledge distillation, these models may also benefit   from the knowledge encoded by language models that are pre-trained on rich sources of texts. The distillation process, however, is challenging due to the modal disparity between textual and speech embedding spaces. This paper studies metric-based distillation to align the embedding space of text and speech with only a small amount of data without modifying the model structure. Since the semantic and granularity gap between text and speech has been omitted in literature, which impairs the distillation, we propose the \textbf{P}rior-informed \textbf{A}daptive knowledge \textbf{D}istillation (PAD) that adaptively leverages text/speech units of variable granularity and prior distributions to achieve better global and local alignments between text and speech pre-trained models. We evaluate on three spoken language understanding benchmarks to show that PAD is more effective in transferring linguistic knowledge than other metric-based distillation approaches.

\end{abstract}

\section{Introduction}
\label{sec:intro}
Recent years witness a great success of pre-trained models in various domains, e.g., natural language~\cite{DBLP:conf/naacl/DevlinCLT19, DBLP:conf/iclr/LanCGGSS20, DBLP:journals/corr/abs-1907-11692, DBLP:journals/jmlr/RaffelSRLNMZLL20, radford2019language} and speech~\cite{DBLP:conf/interspeech/SchneiderBCA19, DBLP:conf/nips/BaevskiZMA20, DBLP:journals/taslp/HsuBTLSM21}. The pre-trained models learn to encode the input into high-level features via self-supervised learning tasks such as masked language modeling (MLM), next sentence prediction, etc., and bring considerable performance gain to the downstream tasks~\cite{DBLP:conf/naacl/DevlinCLT19, DBLP:conf/iclr/LanCGGSS20, firdaus2023multitask}. Spoken Language Understanding (SLU) is a set of downstream tasks in the speech domain intending to learn models to understand the content of speech input.
State-of-the-art SLU models are mostly achieved by finetuning on the speech pre-trained models~\cite{DBLP:conf/interspeech/YangCCLLLLSCLHT21}. 

Despite that the pre-trained speech encoder is powerful in preserving content-based information, it still falls short in encoding linguistic information due to the absence of symbolic representation. To complement the speech pre-training, a speech encoder could benefit from transferring knowledge from language models that are pre-trained on unlabelled texts. However, the knowledge obtained during the pre-training stage of speech and text has a natural gap caused by the modal bias and the difference in pre-training methods~\cite{DBLP:conf/acl/AoWZ0RW0KLZWQ0W22}. This motivates us to study the problem of distilling knowledge from text pre-trained models to speech ones to enhance their performance on understanding tasks. 


We categorize existing work on leveraging text knowledge for end-to-end SLU into two classes: one-tower and two-tower. The one-tower approaches~\cite{DBLP:conf/acl/AoWZ0RW0KLZWQ0W22, sachidananda2022calm} encode uni-modal inputs via a shared encoder and is flexible in taking either uni-modal or multi-modal inputs at inference stage. The two-tower methods use modal-specific encoders for speech and text learning, and rely on either additional layers/codebooks~\cite{DBLP:conf/acl/AoWZ0RW0KLZWQ0W22, bapna2021slam, bapna2022mslam, agrawal2022tie, liu2021cross} or alignment losses to align the multi-modal embedding space. 

This work falls under the two-tower category aligning the text and speech embedding space by designing metric objectives~\cite{DBLP:conf/interspeech/DenisovV20, DBLP:conf/naacl/ChungZZ21}, which we call metric-based distillation. 
Compared with the one-tower methods, metric-based distillation requires only a small amount of data to distill text knowledge~\cite{DBLP:conf/naacl/ChungZZ21} with an explicit alignment objective. It does not introduce any additional parameters, which ensures the model's efficiency and separability. However, due to the large semantic and granularity gap between speech and text, it is challenging to design appropriate metrics for metric-based distillation. Prior works~\cite{DBLP:conf/interspeech/DenisovV20, DBLP:conf/naacl/ChungZZ21} use simple metrics (L2 distance and cosine similarity) to perform metric-based distillation from text to speech ignoring the semantic and granularity difference between them, which harms the distillation effectiveness. To this end, we propose the \textbf{P}rior-informed \textbf{A}daptive knowledge \textbf{D}istillation (PAD) that aligns text-speech representations at adaptive granularity, with the alignment process being informed by the prior knowledge reflecting the semantic significance. 

\vspace{0.5em}
\noindent We made the following \textbf{contributions}:
\vspace{-0.5em}
\begin{itemize}[leftmargin=*]
    \item We propose to apply the Attention-based Significance Priors (ASP) to ease the \textbf{semantic} knowledge transferring from texts to speech. 
    \vspace{-0.5em}
     \item We propose the Anchor-based Adaptive Span Aggregation algorithm (AASA) that narrows the modal \textbf{granularity} gap of alignments.
    \vspace{-0.5em}
    \item To the best of our knowledge, we are the first that evaluate multiple different alignment strategies beyond vanilla global and local alignments to study the feasibility of metric-based speech-text distillations. The results on three spoken language understanding benchmarks verify our assumptions and claims.
\end{itemize}





\section{Methodology}
\label{sec:Methodology}


\subsection{Preliminary: Global and Local Alignment}
\label{subsec:Alignment of different levels}
Metric-based distillation distills text knowledge by aligning the representation space of speech and text pre-trained models. In general, there are two aligning strategies: global and local alignments. Global-level alignment narrows the gap between sequence-level representations and local-level alignment similarizes the local unit representations. Once the alignment training is finished, the speech representation space is closer to that of the text and thus text knowledge is transferred. Given a pair of speech and text data, 
the output of speech and text pre-trained models are denoted as \(\mathrm{\mathbf{S}} = [\mathrm{\mathbf{s}}_1, ..., \mathrm{\mathbf{s}}_n]\) and \(\mathrm{\mathbf{T}} = [\mathrm{\mathbf{t}}_1, ..., \mathrm{\mathbf{t}}_n]\) respectively, where 
\(\mathrm{\mathbf{s}}_m \in \mathbb{R}^{d_{s}}\) and \(\mathrm{\mathbf{t}}_m \in \mathbb{R}^{d_{t}}\). \(d_s\) and \(d_t\) are the model dimension of speech and text modules. 
\vspace{-1em}
\paragraph*{Global-level alignment.} Sequence-level representations can be extracted from the output representations of the pre-trained models. We denote the global-level output representation of speech and text model as \(\mathrm{\mathbf{\hat{s}}}  \in \mathbb{R}^{d_{s}}\) and \(\mathrm{\mathbf{\hat{t}}} \in \mathbb{R}^{d_{t}}\), which are downsampled embeddings that represents the sentence such as the \(cls\) embedding in BERT or the averaged embeddings. The L1 distance of the two sequence representations is computed for alignment: 

\vspace{-1em}
\begin{equation}
\label{equ:Gsen}
\mathcal{L}_{Glob} = ||\mathrm{\mathbf{\hat{s}}} - \mathrm{\mathbf{\hat{t}}}||_1
\end{equation}
\vspace{-1em}

During global-level alignment training, the text module is frozen and only the speech module is updated.
\vspace{-1em}
\paragraph*{Local-level alignment.} A finer alignment can be conducted locally. we cannot use the L1 distance as in Eq.\ref{equ:Gsen} because the speech and text tokens do not have a one-to-one mapping.
The local token units of speech and text module are aligned by maximizing the sum of the maximum text-speech similarities regarding all text units:

\vspace{-1em}
\begin{equation}
\label{equ:Ltok}
\mathcal{L}_{Loc} = -\frac{1}{n} \sum_{j=0}^n max_i \phi(\mathrm{\mathbf{s}}_i, \mathrm{\mathbf{t}}_j)
\end{equation}
\vspace{-1em}

Where \(\phi\) denotes the similarity metric such as cosine similarity. Same as the global-level alignment, the text module is fixed and only the speech module is updated. 

\begin{figure}[t]
\centering
\includegraphics[width=0.47\textwidth]{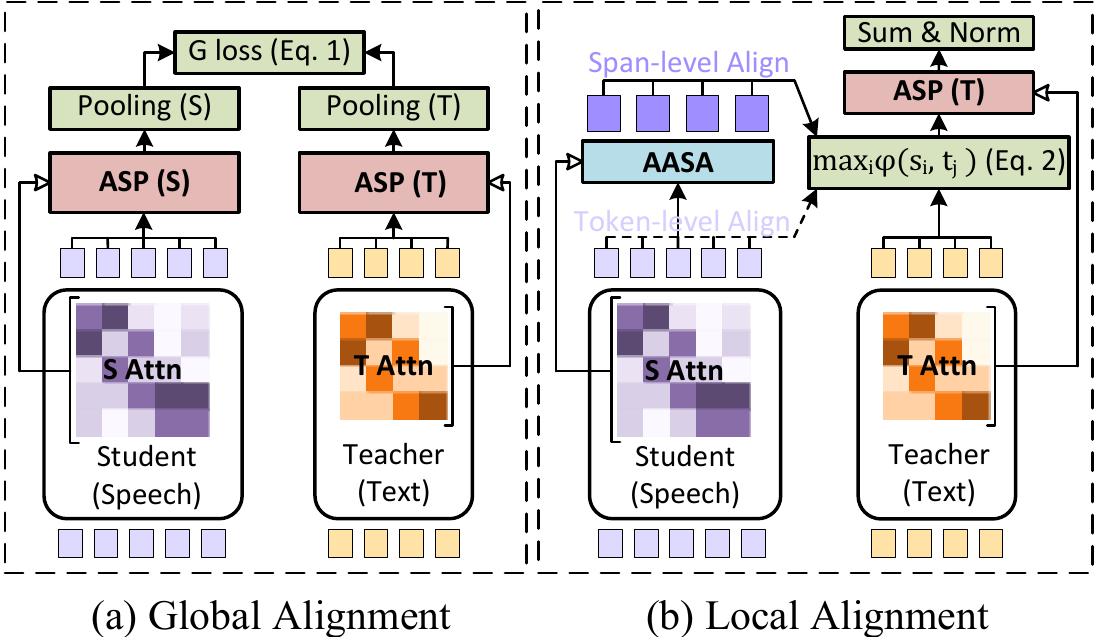}
\caption{The global and local PAD. The global and local alignments (Sec.\ref{subsec:Alignment of different levels}) are both informed by the \textbf{ASP} (Sec.\ref{sec:ASP}) to narrow the semantic gap. The \textbf{AASA} (Sec.\ref{sec:AASA}) adaptively reorganizes the speech sequence to narrow the granularity gap.}
\label{fig:model}
\end{figure}

\subsection{Attention-based Significance Priors}
\label{sec:ASP}
Both global and local alignments introduced in Sec.\ref{subsec:Alignment of different levels} treat tokens in a sequence equally. However, the modal bias of speech and text causes some alignments to be meaningless, e.g., blank and noise signals of acoustic sequences cannot be found in text, and trying to align these signals with text tokens may not be beneficial to model's understanding ability. Thus, we assume that the alignment between acoustic features and their text counterpart should be focused on only the most semantic-relevant tokens or frames, for example, the non-blank speech tokens or the keywords in the text sequence. 
As a result, we need to properly score the significance of tokens (or frames). 
\cite{DBLP:conf/naacl/ChungZZ21} used the inverse document frequency (idf) to re-weight the significance of frequent token-frame pairs. However, idf has two problems: (a) it is fixed for each dataset, whereas the significance of a token varies with different contexts; (b) it is not applicable to speech features that are continuous-valued. We propose Attention-based Significance Priors (ASP) that extracts the significance distribution \(\mathcal{P}_{sig} \in \mathbb{R}^n\) of an output hidden sequence \(\mathrm{\mathbf{H}} = [\mathrm{\mathbf{h}}_1, .. ,\mathrm{\mathbf{h}}_n]\) from its self-attention map of the pre-trained transformer models\footnote{We find that the last-layer attention map achieves a better estimation of the ASP in some cases, which led to better performance.} (Fig.\ref{fig:model}):  

\begin{equation}
\small
\label{eq: prior constrain}
    \mathcal{P}_{sig}(\mathrm{\mathbf{H}}) = \frac{1}{L_0} \sum_{l=1}^{L_0} \frac{1}{\mathrm{\mathbf{e}} A(\mathrm{\mathbf{H}})^l \mathrm{\mathbf{e}}^T} \sum_{m=1}^n A(\mathrm{\mathbf{H}})^l_m
\end{equation}

The significance score of \(\mathrm{\mathbf{h}}_m\) is a normalized mean over the attention weights of the units attending to \(\mathrm{\mathbf{h}}_m\). \(A(\mathrm{\mathbf{H}})^l = [A(\mathrm{\mathbf{H}})^l_1, ..., A(\mathrm{\mathbf{H}})^l_n]\) denotes the \(l\)-th layer self-attention map; \(A(\mathrm{\mathbf{H}})^l_m \in \{a\in \mathbb{R} | 0<a<1\}^n\) denotes the attention weights of the units that attend to the \(m\)-th unit of $\mathrm{\mathbf{H}}$, which is the \(m\)-th row of \(A(\mathrm{\mathbf{H}})^l\); \(L_0\) is the number of layers of the transformer model; $\mathrm{\mathbf{e}} = \{1\}^n$. The prior distribution is contextual and we will show that the priors of both speech and text modules narrow the modal semantic gap and contribute to better alignment results. 

For global alignments, we obtain \(\mathrm{\mathbf{S}}'\) and \(\mathrm{\mathbf{T}}'\) after applying the significance prior to the output sequence of speech and text module: \(\mathrm{\mathbf{S}}' = \mathrm{\mathbf{S}} \odot \mathcal{P}_{sig}(\mathrm{\mathbf{S}})\), \(\mathrm{\mathbf{T}}' = \mathrm{\mathbf{T}} \odot \mathcal{P}_{sig}(\mathrm{\mathbf{T}})\). For local alignments, according to Eq.\ref{equ:Ltok}, we obtain the similarity sequence \(\mathrm{\mathbf{\Phi}} = [\mathrm{\mathbf{\varphi}}_1, ..., \mathrm{\mathbf{\varphi}}_n] \) regarding the text output $\mathrm{\mathbf{T}}$, and we obtain \(\mathrm{\mathbf{\Phi}}'\) constrained by prior: \(\mathrm{\mathbf{\Phi}}' = \mathrm{\mathbf{\Phi}} \odot \mathcal{P}_{sig}(\mathrm{\mathbf{\Phi}})\). For PAD, we replace the original sequences with their prior constrained ones in Eq.\ref{equ:Gsen} and Eq.\ref{equ:Ltok}. 

\subsection{Anchor-based Adaptive Span Aggregation}
\label{sec:AASA}
In reality, a text token may correspond to many speech tokens since a word consists of multiple phonemes. However, the local alignment defined in Sec.\ref{subsec:Alignment of different levels} aligns each text token with the corresponding speech token, which is not consistent in granularity. Such alignment is possible because the attention representations are contextual - each speech embedding potentially represents the contextualized phoneme. However, attention scores are still highly localized for transformers~\cite{DBLP:conf/acl/LiYYHC20, DBLP:conf/iclr/WuLLLH20}, meaning that the embeddings tend to only represent their close neighbors. As a result, there still remains a granularity gap between the representations of speech and text tokens, thus directly aligning them is not optimal. 

\begin{algorithm}[t]
\caption{Anchor-based Adaptive Span Aggregation}
\label{algorithm: Adaptive Span Aggregation}
\begin{algorithmic}[1]
\State \textbf{Procedure} AASA ($\mathrm{\mathbf{S}}, \xi$)
\textcolor{green1}{\Comment{Input: $\mathrm{\mathbf{S}} \in  \mathbb{R}^{n \times d_s}$}}
\State Anchor points \(\mathrm{\mathbf{\Gamma}} \gets [id_{e0}] \,\); span pools \(\tilde{\mathrm{\mathbf{S}}} \gets [ \,\ ] \,\).
\State \(P_{sig}'(\mathrm{\mathbf{S}})\), \(Index_{sorted}\) \(\gets\) Descending Sort \(P_{sig}(\mathrm{\mathbf{S}})\).\\
\textcolor{green1}{\Comment{Select anchor pts by their significance scores (Eq.\ref{eq: prior constrain}).}}
\For {\(id\) in \(Index_{sorted}\)} 
\If {\(\forall id_e \in \mathrm{\mathbf{\Gamma}}, |id - id_{e}| > \xi/2\)}
\textcolor{green1}{\Comment{Control density.}}
\State $span_m \gets \{\mathrm{\mathbf{s}}_{id-s_m}, ..., \mathrm{\mathbf{s}}_{id+s_m}\}$ for $m \in {1, ..., c}$\\
\textcolor{green1}{\Comment{$s_m \in \{\xi/2, \xi, 2\xi, ...\}$; $\xi$ denotes the base scale.}}
\State $\mathrm{\mathbf{\tilde{s}}}_{id} \gets \{(pooling(span_1), ... , pooling(span_c)\}$
\State Add \(id\) to \(\mathrm{\mathbf{\Gamma}}\); add \(\mathrm{\mathbf{\tilde{s}}}_{id}\) to \(\mathrm{\mathbf{\tilde{S}}}\).
\EndIf
\EndFor
\State Return $\mathrm{\mathbf{\tilde{S}}}$ \textcolor{green1}{\Comment{Output: $\tilde{S} \in \mathbb{R}^{k \times c \times d_s}$}}
\end{algorithmic}
\end{algorithm}

We propose to adaptively generate local spans where each span is aggregated by several speech embeddings to narrow the granularity gap. The alignment process is the same as Eq.\ref{equ:Ltok}, only being different in that the original speech sequence is replaced with the local span pools generated by the Anchor-based Adaptive Span Aggregation (AASA) procedure (Algorithm \ref{algorithm: Adaptive Span Aggregation}). We denote the local span pools of a speech sequence $\mathrm{\mathbf{S}}$ as \(\mathrm{\mathbf{\tilde{S}}} = [\mathrm{\mathbf{\tilde{S}}}_1, ... , \mathrm{\mathbf{\tilde{S}}}_k]\), where \(\mathrm{\mathbf{\tilde{S}}}_m \in \mathbb{R}^{c \times d_s}\). Each span pool \(\mathrm{\mathbf{\tilde{S}}}_m\) consists of \(c\) spans of different scales, and the absolute locations of the span pools are decided by the anchor points. In line with local-level alignment (Eq.\ref{equ:Ltok}), each text token aligns with one of the spans \(\mathrm{\mathbf{\tilde{s}}}_m\) in $\mathrm{\mathbf{\tilde{S}}}$, maximizing the sum of the maximum text-span similarities regarding all text tokens.

\section{Results and Analysis}
\label{sec:Results}
\paragraph*{Experimental setup.}
\label{subsec:Experimental Setup}
We use \textit{wav2vec2-base} and \textit{bert-base-uncased} checkpoints from the Hugging Face\footnote{\url{https://huggingface.co}} as the speech and text modules. We choose 768 dimensions for both models and cut down the BERT vocabulary size to 5000. 
Following~\cite{DBLP:conf/naacl/ChungZZ21}, the alignment training uses a 10-hour transcripted data randomly selected from the \textit{train-clean-360} subset of Librispeech 960~\cite{DBLP:conf/icassp/PanayotovCPK15} dataset. 
For the evaluation of all downstream tasks, we follow the SUPERB~\cite{DBLP:conf/interspeech/YangCCLLLLSCLHT21} benchmark using S3PRL\footnote{\url{https://github.com/s3prl/s3prl}} to achieve fair comparison. We evaluate three SLU tasks: Intent Detection (IC), Emotion Recognition (ER), and Slot Filling (SF) on Fluent Speech Commands, IEMOCAP, and Audio SNIPS datasets, respectively. 

SUPERB freezes the upstream pre-trained model and controls all other settings to be identical.~\cite{DBLP:conf/naacl/ChungZZ21} finetune the whole upstream model, and they compare with baselines without specifying the settings to be identical. Thus, their alignment effectiveness is not easily comparable and our results based on SUPERB are fairer and more reproducible for future work. 
We re-implement their alignment methods for comparison.

\begin{table}[t]
\setlength{\tabcolsep}{4.2pt}
\centering
\footnotesize
\caption{Results on IC, ER, and SF in terms of accuracy and F1 score. Our PAD outperforms all the alignment baselines.}
\vspace{-1.2em}
\centering
\begin{tabular}{@{}lcccc@{}}
\toprule
\multirow{2}{*}{Models} & \multirow{2}{*}{\begin{tabular}[c]{@{}c@{}}Align\\ Variant \end{tabular}} & \begin{tabular}[c]{@{}c@{}}Intent   \\ Classification\end{tabular} & \begin{tabular}[c]{@{}c@{}}Emotion   \\ Recognition\end{tabular} & \begin{tabular}[c]{@{}c@{}}Slot   \\ Filling\end{tabular} \\ \cmidrule(l){3-5} 
                                 &  & Acc   $\uparrow$                                      & Acc   $\uparrow$                                    & F1   $\uparrow$                             \\ \midrule
wav2vec 2.0 Base~\cite{DBLP:conf/nips/BaevskiZMA20}       & - & 94.40                         & 63.02                         & 87.94  \\
Glob -   cls~\cite{DBLP:conf/naacl/ChungZZ21}  & G & 97.39                                                              & 64.19                                                            & 86.30                                                     \\
TLocal - idf~\cite{DBLP:conf/naacl/ChungZZ21}                     & T & 94.91                                                              & 58.49                                                            & 82.04                                                     \\ \midrule
Glob -   avr                      & G & 97.30                                                              & 64.65                                                            & 86.00                                                     \\
TLocal                           & T & 95.90                                                              & 59.18                                                            & 82.27                                                     \\

TLocal - CW                       & T & 96.81                                                              & 60.27                                                            & 84.88                                                     \\
TLocal - CG                      & T & 96.76                                                              & 61.97                                                            & 84.89                                                     \\
TLocal - OR                      & T & 90.30                                                              &                                                  59.32           &     77.36                                                      \\
SLocal - OR                    &  S  & 97.57                                                              & 63.87                                                            & 85.12                                                     \\ \midrule
PAD-Glob                       & G  & 97.68                                                              & \textbf{64.91}                                                            & 87.67                                                     \\
PAD-TLocal                   &  T    & 95.90                                                              & 62.46                                                            & \textbf{88.40}                                                     \\
PAD-SLocal                    &  S   & \textbf{97.86}                                                              & 62.74                                                            & 87.19   \\ \bottomrule                                                 
\end{tabular}
\label{tab:baseline1-PAD}
\vspace{-1em}
\end{table}

\paragraph*{Results against various alignment methods.} We report the results of three levels of PAD against the baselines in Tab.\ref{tab:baseline1-PAD}. G, T, and S denote global, token-level local, and span-level local alignment, respectively. For global-level alignment (Eq.\ref{equ:Gsen}), we compare with 2 variants of sentence embeddings: $cls$ embedding (\textbf{Glob-cls})~\cite{DBLP:conf/naacl/ChungZZ21} and averaged pooling  (\textbf{Glob-avr}). For token-level local alignment, we compare with 5 variants: vanilla token-level alignment (\textbf{TLocal}) as in Eq.\ref{equ:Gsen}; token-level alignment using idf to ignore the frequent words (\textbf{TLocal-idf})~\cite{DBLP:conf/naacl/ChungZZ21}; first training a speech token recognizer with CTC loss using the same 10-hour data, then the prediction probabilities on text vocabulary are used to weight (\textbf{TLocal-CW}, \textit{by constraining the alignment as ASP}) or guide (\textbf{TLocal-CG}, \textit{achieve a speech-text mapping by max probability, and align them according to the mapping instead of the max similarity of Eq.\ref{equ:Ltok}}) the alignment; \textbf{TLocal-OR} treats the similarity distribution of Eq.\ref{equ:Ltok} as the probability distribution in CTC calculation and computes the alignment loss as a CTC loss to achieve alignments based on maximum expectations. For span-level local alignment, we compare with the span-level ordered align (\textbf{SLocal-OR}), which is the same as the \textbf{TLocal-OR} except for its granularity. 

The results in Tab.\ref{tab:baseline1-PAD} show that our three alignments outperform all baselines in the downstream tasks. 
Whereas there exist performance trade-offs on different tasks for global and local alignments. The models under global alignments are better at classification tasks (Intent Classification and Emotion Recognition), whereas the ones under local alignments are better at sequence generation tasks (Slot Filling). Classification tasks require better global-level semantics whereas sequence generation tasks require finer token-level semantics. Compared with the token-level alignment, the span-level alignment achieves good performance on both kinds of tasks, which illustrates the effectiveness of AASP that narrows the gap between speech and text granularities. 
\vspace{-1em}

\begin{table}[t]
\setlength{\tabcolsep}{5.5pt}
\footnotesize
\caption{Ablation study on IC, ER, and SF by removing the features one by one.}
\vspace{-1.2em}
\centering
\label{tab:Ablation study}
\begin{tabular}{@{}lcccc@{}}
\toprule
\multirow{2}{*}{Align   Variants} & \multirow{2}{*}{\begin{tabular}[c]{@{}c@{}}Align\\ Variant \end{tabular}}  & \begin{tabular}[c]{@{}c@{}}Intent   \\ Classification\end{tabular} & \begin{tabular}[c]{@{}c@{}}Emotion   \\ Recognition\end{tabular} & \begin{tabular}[c]{@{}c@{}}Slot   \\ Filling\end{tabular} \\ \cmidrule(l){3-5} 
                                 &  & Acc   \(\uparrow\)                                      & Acc   \(\uparrow\)                                    & F1   \(\uparrow\)                              \\ \midrule
\textbf{PAD-Glob}                 & G & \textbf{97.68}                                                     & \textbf{64.91}                                                   & \textbf{87.67}                                            \\
\quad - w/o s   prior                     & G & 97.55                                                              & 64.39                                                            & 87.17                                                     \\
\quad - w/o t   prior                     & G & 97.34                                                              & 64.14                                                            & 87.59                                                     \\
\quad - w/o   both                        & G & 97.30                                                              & 63.97                                                            & 86.29                                                     \\ \midrule
\textbf{PAD-TLocal}              & T & \textbf{95.90}                                                     & \textbf{62.46}                                                   & \textbf{88.40}                                            \\
\quad - w/o   prior                      & T  & 95.90                                                              & 59.18                                                            & 82.27                                                     \\ \midrule
\textbf{PAD-SLocal}              & S  & \textbf{97.86}                                                     & \textbf{62.74}                                                   & \textbf{87.19}                                            \\
\quad - w/o   prior                     & S  & 97.52                                                              & 60.37                                                            & 83.04                                                     \\
\quad - w/o span  pool                   & S  & 97.86                                                              & 62.72                                                            & 87.06                                                     \\
\quad - w/o   anch pts                   & S & 96.73                                                              & 60.88                                                            & 86.10                                                     \\  \bottomrule
\end{tabular}
\vspace{-1em}
\end{table}

\paragraph*{Ablation Study.} We evaluate the impact of the various choices we made for PAD by ablating its features (Tab.\ref{tab:Ablation study}). For PAD-Glob, we remove the speech and text priors one by one and the consistent performance drop verifies their importance. Similarly, we also verify the effectiveness of the significance priors for PAD-TLocal and PAD-SLocal. Additionally, for PAD-SLocal, we find that either (a) \textit{removing the span pools and using fixed spans} or (b) \textit{using even-stride spans instead of anchor-based ones} decreases the performance.

\vspace{-1em}
\paragraph*{Analysis of joint alignments.} We also evaluate all the joint alignment combinations that perform multiple alignments concurrently (Tab.\ref{tab:joint alignment}). We find that all of them perform worse than the combined results of separate alignments. Furthermore, we observe that the performances of PAD-G\&S and PAD-T\&S $>$ PAD-all joint $>$ PAD-G\&T, which is in line with our assumptions: the global-level and token-level alignment may be incompatible in essence because of their granularity gap, whereas the granularity of both of them are close to the span-level alignment because of the adaptive aggregation of AASA (Eq.\ref{algorithm: Adaptive Span Aggregation}).

\begin{table}[t]

\setlength{\tabcolsep}{7pt}
\centering
\footnotesize
\caption{Analysis of the joint alignment combinations.}
\vspace{-1.2em}
\label{tab:joint alignment}
\begin{tabular}{@{}lcccc@{}}
\toprule
\multirow{2}{*}{Align   Variants} & \multirow{2}{*}{\begin{tabular}[c]{@{}c@{}}Align\\ Variant \end{tabular}}  & \begin{tabular}[c]{@{}c@{}}Intent   \\ Classification\end{tabular} & \begin{tabular}[c]{@{}c@{}}Emotion   \\ Recognition\end{tabular} & \begin{tabular}[c]{@{}c@{}}Slot   \\ Filling\end{tabular} \\ \cmidrule(l){3-5} 
                                 &  & Acc   \(\uparrow\)                                      & Acc   \(\uparrow\)                                    & F1   \(\uparrow\)                              \\ \midrule
PAD-Glob                 & G & 97.68                                                     & \textbf{64.91}                                                   & 87.67                                            \\
PAD-TLocal              & T & 95.90                                                     & 62.46                                                   & \textbf{88.40}                                            \\
PAD-SLocal              & S  & \textbf{97.86}                                                     & 62.74                                                   & 87.19                                            \\ \midrule
PAD-G\&T                & G+T  & 68.63                                                              &                                   61.42                               & 77.98                                                     \\
PAD-G\&S                 & G+S & 97.86                                                              &                                   62.93                               & 87.25                                                     \\
PAD-T\&S               & T+S & 97.68                                                              &                                   62.91                               & 86.49                                                     \\
PAD-all   joint                 & G+T+S  & 97.50                                                              &                                   62.40                               & 85.47                                                     \\ \bottomrule
\end{tabular}
\vspace{-1em}
\end{table}

\begin{figure}[h]
\centering
\includegraphics[width=0.49\textwidth]{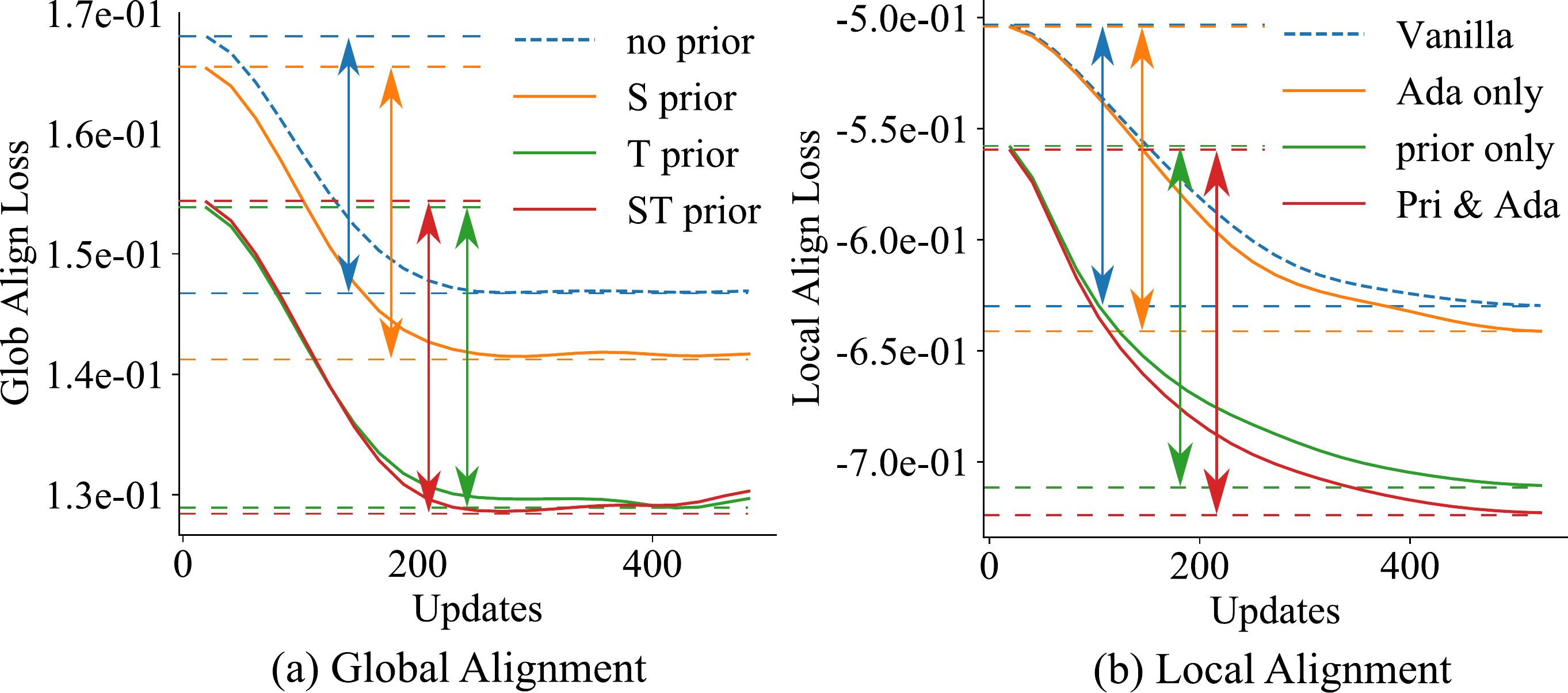}
\caption{The global and local-level alignment loss on dev set. S and T denote speech and text, respectively.}
\label{fig:loss}
\end{figure}

\vspace{-1em}
\paragraph*{Analysis of alignment loss.} Apart from the downstream task accuracies, the variation of alignment loss before and after the alignment is a rough metric for comparing the alignment effectiveness since they reflect the homogenization amount of embedding space. With all entries being properly scaled, we plot the validation loss curves of global and local alignment training from epoch 1 in Fig.\ref{fig:loss}. Fig.\ref{fig:loss}(a) shows that the three alignments constrained by the prior show higher loss deduction compared with the vanilla version, which is in line with the results obtained above, illustrating the effectiveness of the speech and text priors. In Fig.\ref{fig:loss}(b), the local alignments that are constrained by the priors or leverage the adaptive spans show higher loss deduction, which is also in line with the effectiveness of the significance prior and adaptive span in local-level alignment experiments. 

\vspace{-1em}

\section{Conclusion}
\label{sec:Conclusion}

We propose a method that leverages adaptive spans and the in-stored significance prior distribution to achieve better alignments between text and speech pre-trained models at both global and local levels. 
We also observe that the trade-off between global and local alignments always exists because they may be contradictory in essence. We plan to investigate how to alleviate this trade-off in future work.


\newpage
\bibliographystyle{IEEEbib}
\bibliography{strings,refs}

\end{document}